\documentclass{article}

\usepackage{spconf, amsmath,graphicx,cases,multirow,indentfirst,subeqnarray,amsfonts,amssymb,subfigure }

\title{UNSUPERVISED DOMAIN ADAPTION DICTIONARY LEARNING FOR VISUAL RECOGNITION}
\name{Zhun Zhong \qquad Zongmin Li \qquad Runlin Li \qquad Xiaoxia Sun}

\address{College of Computer \(\&\) Communication Engineering, China University of Petroleum, Qingdao, China\\zhun.zhong@foxmail.com, lizongmin@upc.edu.cn,   \{rlli,xxsun\}@s.upc.edu.cn}

\begin{document}
%
\maketitle

\begin{abstract}


Over the last years, dictionary learning method has been extensively applied to deal with various computer vision recognition applications, and produced state-of-the-art results. However, when the data instances of a target domain have a different distribution than that of a source domain, the dictionary learning method may fail to perform well. In this paper, we address the cross-domain visual recognition problem and propose a simple but effective unsupervised domain adaption approach, where labeled data are only from source domain.  In order to bring the original data in source and target domain into the same distribution, the proposed method forcing nearest coupled data  between source and target domain to have identical sparse representations while jointly learning dictionaries for each domain, where the learned dictionaries can reconstruct original data in source and target domain respectively. So that sparse representations of original data can be used to perform visual recognition tasks. We demonstrate the effectiveness of our approach on standard datasets. Our method performs on par or better than competitive state-of-the-art methods.

\end{abstract}
\section{Introduction}
\label{sec:intro}

In the past decade, machine learning has been widely used for various computer vision applications, such as object recognition \cite{zhang2011sparse}, multimedia retrieval  \cite{zheng2015fast,zheng2014coupled,kuang2015retrieval}, image classification \cite{sanchez2013image}, etc. Traditional machine learning methods often learn a model from the training data, and then apply it to the testing data. The fundamental assumption here is that the training data and testing data have the same distribution. However, in real-world applications, it cannot always guarantee that training data share the same distribution with testing data. Therefore, it may produce very poor results when the testing data and training data have the different distributions since the training model is no longer optimal on testing data. For example, applies image classification classifier trained on amazon dataset to phone photos in real life. Face recognition model trained on frontal and well-illumination images to recognize non-frontal poses and less-illumination images. This often viewed as visual domain adaption problem which has been increasing interest in understanding and overcoming.

  Domain Adaption aims at learning an adaptive classifier by utilizing the information between source domain with a plenty of labeled data and target domain which is collected from a different distribution. Generally, we can divide domain adaption into two settings depending on the availability of labels in the target domain data: semi-supervised domain adaption, and unsupervised domain adaption. 
  In scenario of semi-supervised domain adaption, labeled data is available in both source domain (with a plenty of labeled data) and target domain (with a few labeled data), while in scenario of unsupervised domain adaptation labeled data are only available from source domain.  In this paper, we mainly focus on unsupervised domain adaptation which is a more challenging task, and more in line with the real-world applications.

Many recent works \cite{gopalan2011domain,gong2012geodesic,li2009maximizing} focus on subspace based method to tackle visual domain adaption problems. In \cite{li2009maximizing}, Li et al. determined a feature subspace via canonical correlation analysis (CCA) \cite{hotelling1936relations} for recognizing faces with different poses. In \cite{gopalan2011domain}, Gopalan et al. using geodesic flows to generate intermediate subspaces along the geodesic path between source domain subspace and target domain subspace on the Grassmann manifold. In \cite{gong2012geodesic}, Gong et al. proposed Geodesic Flow Kernel (GFK), which computes a symmetric kernel between source and target points based on geodesic flow along a latent manifold. 

\begin{figure*}[t]
\label{fig:fig1}
\centering {\includegraphics[width=17cm,height=4cm]{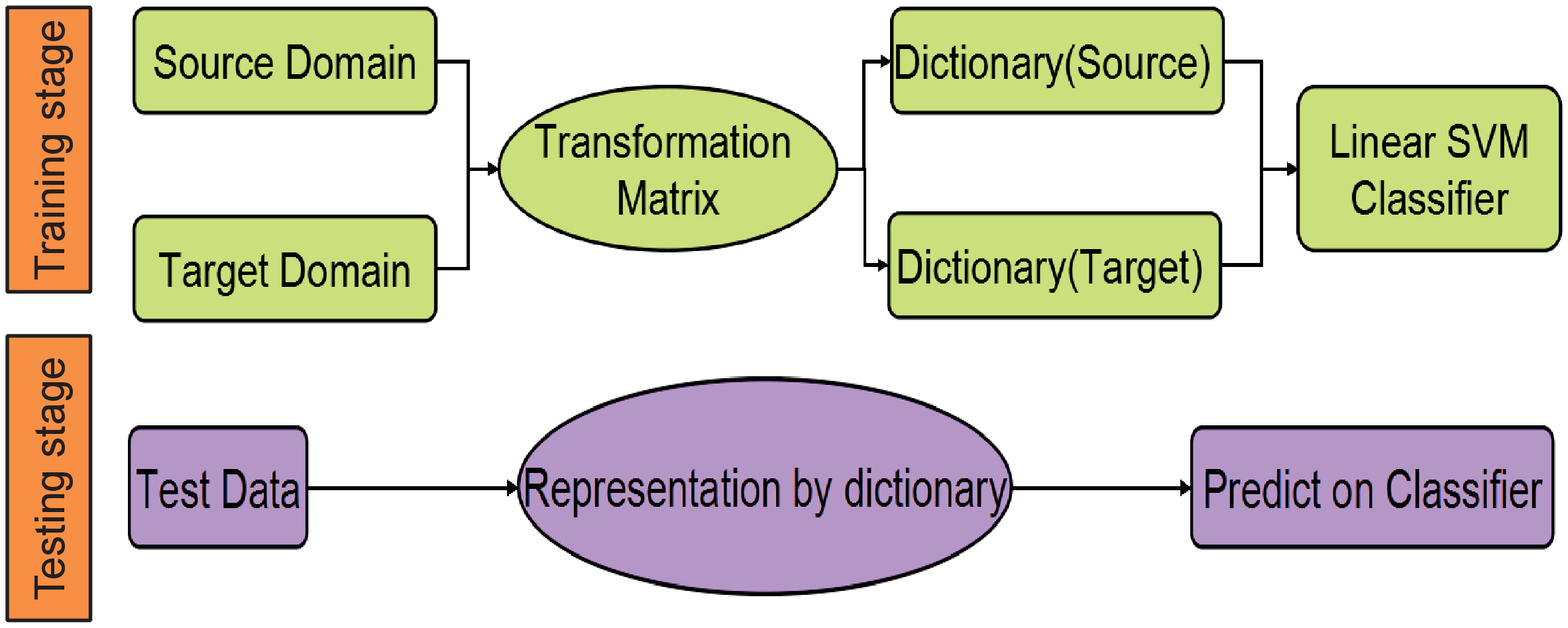}}
             \caption  {The overall schema of the proposed framework.}
             \end{figure*}

In last few years, the study of dictionary learning based sparse representation has received extensive attention. It has been successfully used for a variety of computer vision applications. For example, classification \cite{huang2006sparse}, recognition \cite{wright2009robust} and denoising \cite{elad2006image}.
 Using an over-complete dictionary, signal or image can be approximated by the combination of only a few number of atoms, that are chosen from the learned dictionary. One of the early dictionary learning algorithms was proposed by Olshausen and Field \cite{olshausen1997sparse}, where a maximum likelihood (ML) learning method was used to sparsely encode images upon a redundant dictionary. Based on the same ML objective function as in \cite{olshausen1997sparse}, Engan et al. \cite{engan1999method} developed a more efficient algorithm, called the method of optimal directions (MOD), in which a closed-form solution for the dictionary update has been proposed. More recently, in \cite{aharon2006svd}, Aharon, Elad and Bruckstein proposed the K-SVD algorithm by generalizing k-means clustering and efficiently learns an over-complete dictionary from a set of training signals. This method has been implemented in a variety of image processing problems. 

The most existing dictionary based methods assuming that training data and testing data come from the same distribution. However, the learned dictionary may not be optimal if the testing data has different distribution from the data used for training. Learning dictionaries under different domain is a challenging task, and gradually become a hot research over the last few years. In \cite{jia2010factorized}, Jia et al. considered a special case where corresponding samples from each domain were available, and learn a dictionary for each domain. Qiu et al. \cite{qiu2012domain} presented a general joint optimization function that transforms a dictionary learned from one domain to the other, and applied such a framework to applications such as pose alignment, pose illumination estimation, and face recognition. Zheng et al. \cite{zheng2012cross} proposed a method achieved promising results on the cross-view action recognition problem with pairwise dictionaries constructed using correspondences between the target view and the source view. In \cite{shekhar2013generalized}, Shekhar et al. learn a latent dictionary which can succinctly represent both the domains in a common projected low-dimensional space. Ni et al. \cite{ni2013subspace} learn a set of subspaces through dictionary learning to mitigate the divergence of source and target domains. Huang and Wang \cite{huang2013coupled} proposed a joint model which learns a pair of dictionaries with a feature space for describing and associating cross-domain data. In \cite{zhu2013enhancing,zhu2014weakly}, Zhu and Shao proposed a weakly-supervised framework learns a pairwise dictionaries and a classifier while considering the capacity of the dictionaries in terms of reconstructability, discriminability and domain adaptability.

In this paper, we present an unsupervised domain adaption approach through dictionary learning. Different from above dictionary learning based domain adaption methods, our method directly learning adaptive dictionaries in low-level feature space and with no need for labels either in source domain or target domain during dictionary learning process. Our method is inspired by \cite{zhu2013enhancing,zhu2014weakly}, which forcing the similar samples in the same class to have identical representations in the sparse space. However, our method is unsupervised, we assume that the nearest coupled low-level features in the original space should maintain their relationship in the sparse space (i.e. these coupled features have the same sparse representation).  According to this main idea, we learn a transformation matrix, which selected the nearest data in source domain to each target data. Then the dictionaries for each domain are jointly learned by these selected source data and target data. The data from each domain can be encoded by their dictionaries and then represented by sparse features. Thus, SVM classifier can be trained using these sparse features, and predicting test data on the learned classifier. The learning framework is performed by a classic and efficient dictionary learning method, K-SVD \cite{aharon2006svd}. We demonstrate the effectiveness of our approach on standard cross-domain datasets, and we get state-of-the-art results.  An overall schema of the proposed framework is shown in Fig.1.



\section{PROPOSED METHOD}
\label{sec:2}

\subsection{Problem Notation}
\label{ssec:2.1}
Let 
\(I_{s} = \{I_{s,i}\}_{i=1}^{N_s}\), and
\(I_t = \{I_{t,j}\}_{j=1}^{N_t}\) be the data instances from the source and target domain respectively, where
\(N_s\) and 
\(N_t\) denote the number of samples. Each sample from
\(I_s\) and
\(I_t\) 
has a set of
\(d\)-dimensional local features, thus each sample can represented by
\(I_{s,i}=\{I_{s,i}^1,I_{s,i}^2,...,I_{s,i}^{M_i}\}\)
and
\(I_{t,j}=\{I_{t,j}^1,I_{t,j}^2,...,I_{t,j}^{M_j}\}\)
in source and target domain respectively, where
\(M_i\)
and
\(M_j\)
denote the number of local features. Then, the set of local features of source and target domain can be denoted as
\(Y_s\in\mathbb{R}^{d*L_s}\), and \(Y_t\in\mathbb{R}^{d*L_t}\) respectively, where
\(L_s\) and 
\(L_t\) denote the number of local features in the source and target domain.

\subsection{Dictionary Learning}
\label{ssec:2.2}
Here, we give a brief review of classical dictionary learning approach. Given a set of 
\(d\)-dimensional input signals, 
\(Y\in\mathbb{R}^{d*L}\), where
\(L\) is denoted as the number of input signals. Then, learning a 
\(K\)-atoms dictionary of the signals 
\(Y\), 
\(D\in\mathbb{R}^{d*K}\), can be obtained by solving the following optimization problem:
\begin{flalign}
\begin{split}
\{D,X\}=arg min_{D,X}\|Y-DX\|^2_F  \\
s.t. \text{ }\text{ }\forall_i, \|x_i\|_0 \le T_0
\end{split}
\end{flalign}
where 
\(D=[d_1,d_2,...,d_K]\in\mathbb{R}^{d*K}\) denotes the dictionary, 
\(X=[x_1,x_2,...,x_L]\in\mathbb{R}^{K*L}\) denotes the sparse coefficients of  
\(Y\) decomposed with
\(D\), and
\(T_0\) is the sparsity level that constraint the number of nonzero entries in
\(x_i\).

The performance of sparse representation strictly lie on dictionary learning method. The K-SVD algorithm \cite{aharon2006svd} is a highly effective dictionary learning method that focuses on minimizing the reconstruction error.  In this paper, we will solve our formulation of unsupervised domain adaption dictionary learning based on the K-SVD algorithm.

\subsection{Unsupervised Domain Adaption Dictionary Learning}
\label{ssec:2.3}
Now, consider a more general scenario, where we have data from two domains, source domain 
\(Y_s\in\mathbb{R}^{d*L_s}\), and target domain 
\(Y_t\in\mathbb{R}^{d*L_t}\). We wish to jointly learning corresponding dictionaries for each domain. Formally, we desire to minimize the following cost function:
\begin{flalign}
\begin{split}
&\{D_s,D_t,X_s,X_t\} \\
&=arg min_{D_s,D_t,X_s,X_t}\|Y_s - D_s X_s\|^2_F  \\ 
&+ \|Y_t - D_t X_t\|^2_F \text{ } \text{ } \text{ }
s.t. \text{ } \text{ }\forall_i, [\|x_i^s\|_0, \|x_i^t\|_0] \le T_0
\end{split}
\end{flalign}

In addition, in order to maintain the relationship in original feature space, we assume that the nearest coupled low-level features in the original space should also be the nearest couple in the sparse space. Now the new cost function is given by:
\begin{equation}
\begin{split}
&\{D_s,D_t,X_s,X_t\}\\
&=arg min_{D_s,D_t,X_s,X_t}\|Y_s - D_s X_s\|^2_F \\
& + \|Y_t - D_t X_t\|
^2_F + C([X_s X_t]) \\
&s.t. \text{ }\text{ } \forall_i, [\|x_i^s\|_0 ,  \|x_i^t\|_0] \le T_0
\end{split}
\end{equation}
where 
\(D_s=[d_1^s,d_2^s,...,d_K^s]\in\mathbb{R}^{d*K}\) is the learned source domain dictionary, 
\(X_s=[x_1^s,x_2^s,...,x_{L_s}^s]\in\mathbb{R}^{K*L_s}\) is the sparse coefficients of source domain, 
\(D_t=[d_1^t,d_2^t,...,d_K^t]\in\mathbb{R}^{d*K}\) is the learned target domain dictionary, and 
\(X_t=[x_1^t,x_2^t,...,x_{L_t}^t]\in\mathbb{R}^{K*L_t}\) is the sparse coefficients of target domain. The function 
\(C(\cdot)\) is defined as the distance in the new sparse space of original nearest couples, a small \(C(\cdot)\) indicates the data maintain more relationship in new sparse space. This idea is inspired by \cite{zhu2013enhancing,zhu2014weakly}, in their method, this function is designed to measure the distances of similar cross-domain instances of the same class. However, our method is exactly unsupervised and directly perform on low-level feature.  Thus, the function 
\(C([X_s X_t])\) is defined as:
\begin{equation}
C([X_s X_t])=\|X_t-X_s P\|_F^2
\end{equation}
where 
\(P\in\mathbb{R}^{L_s*L_t}\) is the transformation matrix which records the nearest couples between the original data in source and target domain, 
\(P\) can be represented by:

\begin{equation}       
P=\left(                 
  \begin{array}{cccc}   
    \Phi(y_1^s,y_1^t) & ... & ... & \Phi(y_1^s,y_{L_t}^t)\\  
    \vdots & \ddots & & \vdots \\
    \vdots & & \ddots &  \vdots \\
    \Phi(y_{L_s}^s,y_1^t) & ... & ... & \Phi(y_{L_s}^s,y_{L_t}^t)\\  
  \end{array}
\right)                 
\end{equation}
where 
$\Phi(y_i^s,y_j^t)$ is the Gaussian distance between data in original feature space:

\begin{equation}       
\Phi(y_i^s,y_j^t)= \frac{1}{\sqrt{2\pi}}e^{(-\frac{{y_i^s}^2-{y_j^t}^2}{2})}
\end{equation}
Then, $P$ can be computed by selecting the maximum entry in each column and set to 1 while the other entries are set to 0:
\begin{equation}       
P=(i,j)=\begin{cases}
1& if \quad P(i,j)=max (P(:,j))\\
0& \text{otherwise.}
\end{cases}
\end{equation}
Thus, Equation (3) can be written as:
\begin{flalign}
\begin{split}
&\{D_s,D_t,X_s,X_t\}\\
&=arg min_{D_s,D_t,X_s,X_t}\|Y_s - D_s X_s\|^2_F \\
&+ \|Y_t - D_t X_t\|
^2_F + \|X_t-X_sP\|_F^2 \\
&s.t. \text{ }\text{ }\forall_i, [\|x_i^s\|_0 \|x_i^t\|_0] \le T_0
\end{split}
\end{flalign}
Assuming 
$P$ leads to a perfect mapping across the sparse codes 
$X_t$ and 
$X_s$, and each nearest couple has an identical representation after encoding, then 
$\|X_t-X_sP\|_F^2=0$. 
Thus 
$X_t=X_sP$, we can rewritten Equation (8) as:
\begin{equation}
\begin{split}
&\{D_s,D_t,X_s,X_t\}  \\
&=arg min_{D_s,D_t,X_s,X_t}\|(Y_s - D_s X_s)P\|^2_F + \|Y_t - D_t 
X_t\|^2_F \\
&=arg min_{D_s,D_t,X_s,X_t}\|Y_sP - D_s X_sP\|^2_F + \|Y_t - D_t 
X_t\|^2_F \\
&=arg min_{D_s,D_t,X_s,X_t}\|Y_sP - D_s X_t\|^2_F + \|Y_t - D_t X_t
\|^2_F \\
 & s.t. \text{ }\text{ } \forall_i, \|x_i^t\|_0 \le T_0 \\
\end{split}
\end{equation}

\subsection{Optimization}
\label{ssec:2.4}
We can written Equation (9) as:
\begin{flalign}
\begin{split}
\{\widetilde{D},\widetilde{X}\}=arg min_{\widetilde{D},\widetilde{X}}\|\widetilde{Y}-\widetilde{D}\widetilde
{X}\|^2_F \\
s.t. \text{ }\text{ }\text{ } \forall_i, \|\widetilde{x}_i\|_0 \le T_0
\end{split}
\end{flalign}
where 
$\widetilde{Y}=\left
(\begin{matrix}
     Y_sP  \\
     Y_t 
     
\end{matrix}
\right)
$
, 
$\widetilde{D}=\left
(\begin{matrix}
     D_s  \\
     D_t 
     
\end{matrix}
\right)
$
,and 
$\widetilde{X}=X_t
$.
 Thus, such optimization problem can be solved using the K-SVD algorithm \cite{aharon2006svd}.
 
 \subsection{Object recognition}
\label{ssec:2.5}
Given the learned
$D_s$ and 
$D_t$, we obtain sparse representations of the training data in source domain and testing data in target domain respectively. For each image, we obtain a set of sparse representation 
\(X_i=[x_{i,1},x_{i,2},...,x_{i,M_i}]\in\mathbb{R}^{K*M_i}\), where 
$X_{i,j}$ is the sparse representation of 
$j^{th}$
feature in image 
$i$, 
$K$ denotes the dictionary size, and 
$M_i$ is the number of local feature in image
$i$. Then each image represented by a 
$K$-vector global representation through max pooling the sparse codes of local features, and then we use linear SVM classifier for cross-domain recognition.

%
%
%
%

\section{EXPERIMENTS}
\label{sec:3}

In this section, we evaluate our domain adaption ap-proach on 2D object recognition across different datasets.

\textbf{Experimental Setup:} Following the experiment setting in \cite{gong2012geodesic}, we evaluate our domain adaption approach on four datasets: Amazon (images downloaded from online mer-chants), Webcam (low resolution images by a web camera), Dslr (high-resolution images by a SLR camera), and Caltech-256 \cite{griffin2007caltech}. We regard each dataset as a domain. 
 We extract 10 classes common to all four datasets: BACKPACK, TOURING-BIKE, CALCULATOR, HEADPHONES, COMPUTER-KEYBOARD, LAPTOP-101, COM-  PUTER-MONITOR, COMPUTER-MOUSE, COFFEEMUG, AND VIDEO-PROJECTOR. There are 2533 images in total. Each class has 8 to 151 images in a dataset. We use a SURF detector \cite{bay2008speeded} to extract local features over all images. For each pair of source and target domains, we use 20 training samples per class for Amazon/Caltech, and 8 samples per class for DSLR/Webcam when used as source. To draw complete comparison with existing domain adaption methods, we also carried out experiments on the semi-supervised setting where we additionally sampled 3 labeled images per class from the target domain. We ran 20 different trials corresponding to different selections of labeled data from the source and target domains and testing all unlabeled data in target domain. Our baseline is BOW, where all the images were represented by 800-bin histograms over the codebooks trained from a subset of Amazon images. Our method is also compared with Metric \cite{saenko2010adapting}, SGF \cite{gopalan2011domain} and GFK \cite{gong2012geodesic}. Note that, Metric \cite{saenko2010adapting} is limited to the semi-supervised setting.

\begin{figure}[t]
\label{fig:fig3}
\centering \scalebox{0.42}{\includegraphics{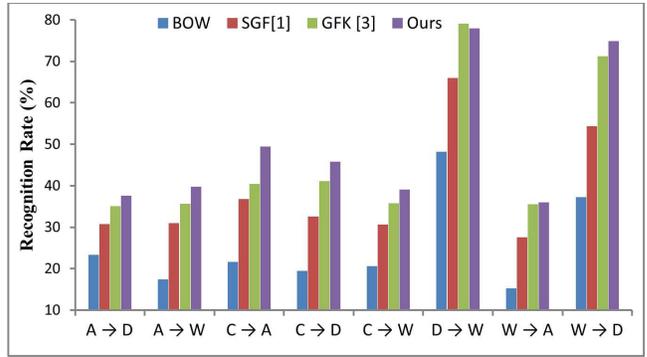}}
             \caption{Cross dataset object recognition accuracies on target domains with unsupervised adaptation over the four datasets (A: Amazon, C: Caltech, D: Dslr, W: Webcam).}
\end{figure}
             
\begin{figure}[t]
\label{fig:fig4}
\centering \scalebox{0.42}{\includegraphics{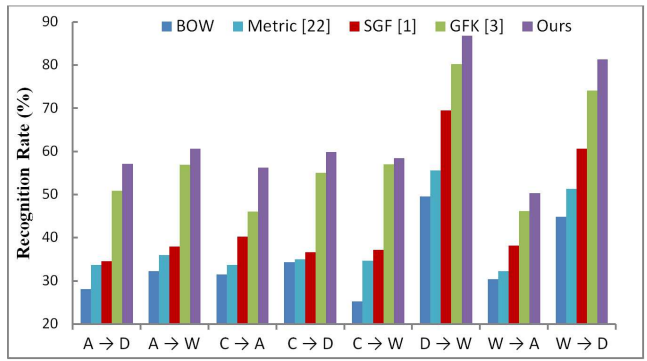}}
             \caption{Cross dataset object recognition accuracies on target domains with semi-supervised adaptation over the four datasets (A: Amazon, C: Caltech, D: Dslr, W: Webcam).}
\end{figure}

\textbf{Parameter Settings:} For our method, we set dictionary size $K=512$, and sparse level $T_0=5$ for each domain.

\textbf{Results:} The average recognition rate is reported in Figure 3 and Figure 4 for unsupervised and supervised settings respectively. It is seen that the baseline BOW has the lowest recognition rate, all domain adaptation methods improve accuracy over it.  Furthermore, GFK \cite{gong2012geodesic} based method clearly outperforms Metric \cite{saenko2010adapting} and SGF \cite{gopalan2011domain}. Overall, our method consistently demonstrates better performance over all methods except for one pair of source and target combination a little less than GFK \cite{gong2012geodesic} in the unsupervised setting.
\section{CONCLUSIONS}
\label{sec:4}

In this paper, we presented a fully unsupervised domain adaption dictionary learning method to jointly learning domain dictionaries by capturing the relationship between the source and target domain in the original data space. We evaluated our method on publicly available datasets and obtain improved performance upon the state of the art.
\section{acknowledgement}
\label{sec:5}

This work is partly supported by National Natural Science Foundation of China (61379106), the Shandong Provincial Natural Science Foundation (ZR2009GL014, and ZR2013FM-036), the Open Project Program of the State Key Lab of CAD \(\&\) CG (Grant No. A1315), Zhejiang University, the Fundamental Research Funds for the Central Universities (12CX06083A, 12CX06086A, 13CX06007A, and 14CX06010A, 14CX06012A, 15CX06017A).


\label{sec:refs}

\bibliographystyle{IEEEbib}
\bibliography{refs}

\begin{thebibliography}{10}

\bibitem{zhang2011sparse}
D~Zhang, Meng Yang, and Xiangchu Feng,
\newblock ``Sparse representation or collaborative representation: Which helps
  face recognition?,''
\newblock in {\em ICCV}. IEEE, 2011, pp. 471--478.

\bibitem{zheng2015fast}
Liang Zheng, Shengjin Wang, Ziqiong Liu, and Qi~Tian,
\newblock ``Fast image retrieval: Query pruning and early termination,''
\newblock {\em Multimedia, IEEE Transactions on}, vol. 17, no. 5, pp. 648--659,
  2015.

\bibitem{zheng2014coupled}
Liang Zheng, Shengjin Wang, and Qi~Tian,
\newblock ``Coupled binary embedding for large-scale image retrieval,''
\newblock {\em Image Processing, IEEE Transactions on}, vol. 23, no. 8, pp.
  3368--3380, 2014.

\bibitem{kuang2015retrieval}
Zhenzhong Kuang, Zongmin Li, Xiaxia Jiang, Yujie Liu, and Hua Li,
\newblock ``Retrieval of non-rigid 3d shapes from multiple aspects,''
\newblock {\em Computer-Aided Design}, vol. 58, pp. 13--23, 2015.

\bibitem{sanchez2013image}
Jorge S{\'a}nchez, Florent Perronnin, Thomas Mensink, and Jakob Verbeek,
\newblock ``Image classification with the fisher vector: Theory and practice,''
\newblock {\em International journal of computer vision}, vol. 105, no. 3, pp.
  222--245, 2013.

\bibitem{gopalan2011domain}
Raghuraman Gopalan, Ruonan Li, and Rama Chellappa,
\newblock ``Domain adaptation for object recognition: An unsupervised
  approach,''
\newblock in {\em ICCV}. IEEE, 2011, pp. 999--1006.

\bibitem{gong2012geodesic}
Boqing Gong, Yuan Shi, Fei Sha, and Kristen Grauman,
\newblock ``Geodesic flow kernel for unsupervised domain adaptation,''
\newblock in {\em CVPR}. IEEE, 2012, pp. 2066--2073.

\bibitem{li2009maximizing}
Annan Li, Shiguang Shan, Xilin Chen, and Wen Gao,
\newblock ``Maximizing intra-individual correlations for face recognition
  across pose differences,''
\newblock in {\em CVPR}, 2009.

\bibitem{hotelling1936relations}
Harold Hotelling,
\newblock ``Relations between two sets of variates,''
\newblock {\em Biometrika}, pp. 321--377, 1936.

\bibitem{huang2006sparse}
Ke~Huang and Selin Aviyente,
\newblock ``Sparse representation for signal classification,''
\newblock in {\em Advances in neural information processing systems}, 2006, pp.
  609--616.

\bibitem{wright2009robust}
John Wright, Allen~Y Yang, Arvind Ganesh, Shankar~S Sastry, and Yi~Ma,
\newblock ``Robust face recognition via sparse representation,''
\newblock {\em Pattern Analysis and Machine Intelligence, IEEE Transactions
  on}, vol. 31, no. 2, pp. 210--227, 2009.

\bibitem{elad2006image}
Michael Elad and Michal Aharon,
\newblock ``Image denoising via sparse and redundant representations over
  learned dictionaries,''
\newblock {\em Image Processing, IEEE Transactions on}, vol. 15, no. 12, pp.
  3736--3745, 2006.

\bibitem{olshausen1997sparse}
Bruno~A Olshausen and David~J Field,
\newblock ``Sparse coding with an overcomplete basis set: A strategy employed
  by v1?,''
\newblock {\em Vision research}, vol. 37, no. 23, pp. 3311--3325, 1997.

\bibitem{engan1999method}
Kjersti Engan, Sven~Ole Aase, and J~Hakon~Husoy,
\newblock ``Method of optimal directions for frame design,''
\newblock in {\em Acoustics, Speech, and Signal Processing}. IEEE, 1999,
  vol.~5, pp. 2443--2446.

\bibitem{aharon2006svd}
Michal Aharon, Michael Elad, and Alfred Bruckstein,
\newblock ``K-svd: An algorithm for designing overcomplete dictionaries for
  sparse representation,''
\newblock {\em Signal Processing, IEEE Transactions on}, vol. 54, no. 11, pp.
  4311--4322, 2006.

\bibitem{jia2010factorized}
Yangqing Jia, Mathieu Salzmann, and Trevor Darrell,
\newblock ``Factorized latent spaces with structured sparsity,''
\newblock in {\em Advances in Neural Information Processing Systems}, 2010, pp.
  982--990.

\bibitem{qiu2012domain}
Qiang Qiu, Vishal~M Patel, Pavan Turaga, and Rama Chellappa,
\newblock ``Domain adaptive dictionary learning,''
\newblock in {\em ECCV}, pp. 631--645. Springer, 2012.

\bibitem{zheng2012cross}
Jingjing Zheng, Zhuolin Jiang, P~Jonathon Phillips, and Rama Chellappa,
\newblock ``Cross-view action recognition via a transferable dictionary
  pair.,''
\newblock in {\em BMVC}, 2012.

\bibitem{shekhar2013generalized}
Sumit Shekhar, Vishal~M Patel, Hien~V Nguyen, and Rama Chellappa,
\newblock ``Generalized domain-adaptive dictionaries,''
\newblock in {\em CVPR}. IEEE, 2013, pp. 361--368.

\bibitem{ni2013subspace}
Jie Ni, Qiang Qiu, and Rama Chellappa,
\newblock ``Subspace interpolation via dictionary learning for unsupervised
  domain adaptation,''
\newblock in {\em CVPR}. IEEE, 2013, pp. 692--699.

\bibitem{huang2013coupled}
De-An Huang and Yu-Chiang~Frank Wang,
\newblock ``Coupled dictionary and feature space learning with applications to
  cross-domain image synthesis and recognition,''
\newblock in {\em ICCV}. IEEE, 2013, pp. 2496--2503.

\bibitem{zhu2013enhancing}
Fan Zhu and Ling Shao,
\newblock ``Enhancing action recognition by cross-domain dictionary learning,''
\newblock in {\em BMVC}, 2013.

\bibitem{zhu2014weakly}
Fan Zhu and Ling Shao,
\newblock ``Weakly-supervised cross-domain dictionary learning for visual
  recognition,''
\newblock {\em International Journal of Computer Vision}, vol. 109, no. 1-2,
  pp. 42--59, 2014.

\bibitem{griffin2007caltech}
Gregory Griffin, Alex Holub, and Pietro Perona,
\newblock ``Caltech-256 object category dataset,''
\newblock 2007.

\bibitem{bay2008speeded}
Herbert Bay, Andreas Ess, Tinne Tuytelaars, and Luc Van~Gool,
\newblock ``Speeded-up robust features (surf),''
\newblock {\em Computer vision and image understanding}, vol. 110, no. 3, pp.
  346--359, 2008.

\bibitem{saenko2010adapting}
Kate Saenko, Brian Kulis, Mario Fritz, and Trevor Darrell,
\newblock ``Adapting visual category models to new domains,''
\newblock in {\em ECCV}, pp. 213--226. Springer, 2010.

\end{thebibliography}

\end{document}